\pdfoutput=1

\documentclass[11pt]{article}

\usepackage[]{EMNLP2022}

\usepackage{times}
\usepackage{latexsym}

\usepackage[T1]{fontenc}

\usepackage[utf8]{inputenc}

\usepackage{microtype}

\usepackage{inconsolata}

\usepackage{tablefootnote}
\usepackage{indentfirst}
\usepackage{array}
\usepackage{graphicx}
\usepackage{adjustbox}
\usepackage{amsmath,amssymb,amsthm,bm}
\usepackage{caption}
\usepackage{multirow}
\usepackage{booktabs}
\usepackage{listings}
\usepackage{subcaption}
\usepackage{paralist}
\usepackage{tablefootnote}
\newcolumntype{P}[1]{>{\centering\arraybackslash}p{#1}}

\usepackage{xspace}
\def\NAME{{IC-DST}\xspace}

\title{In-Context Learning for Few-Shot Dialogue State Tracking}

\author{
     Yushi Hu$^{\spadesuit}$ \quad
     Chia-Hsuan Lee$^{\spadesuit}$\quad
     Tianbao Xie$^{\diamondsuit}$\quad\\
     \bf{
     Tao Yu$^{\diamondsuit\spadesuit}$\quad
     Noah A. Smith$^{\spadesuit\dagger}$\quad
     Mari Ostendorf$^{\spadesuit}$\quad}\\
     $^\spadesuit$University of Washington\\
     $^\diamondsuit$University of Hong Kong\\
     $^\dagger$Allen Institute for Artificial Intelligence\\
     {\tt \{yushihu,chiahlee,ostendor\}@washington.edu}
}

\begin{document}
\maketitle
\begin{abstract}
Collecting and annotating task-oriented dialogues is time-consuming and costly; thus, zero and few shot learning 
could greatly benefit dialogue state tracking (DST).
In this work, we propose an in-context learning (ICL) framework for zero-shot and few-shot learning DST,
where a large pre-trained language model (LM) takes a test instance and a few exemplars as input, and directly decodes the dialogue state \textit{without any parameter updates}.
To better leverage a tabular domain description in the LM prompt, we reformulate DST into a text-to-SQL problem.
We also propose a novel approach to retrieve annotated dialogues as exemplars.
Empirical results on MultiWOZ show that our method IC-DST substantially outperforms previous fine-tuned state-of-the-art models in few-shot settings. In addition, we test \NAME in zero-shot settings, in which the model only takes a fixed task instruction as input, finding that it outperforms previous zero-shot methods by a large margin.\footnote{Our code :
\url{https://github.com/Yushi-Hu/IC-DST}
}
\end{abstract}

\begin{figure}[t]
  \includegraphics[width=0.49\textwidth]{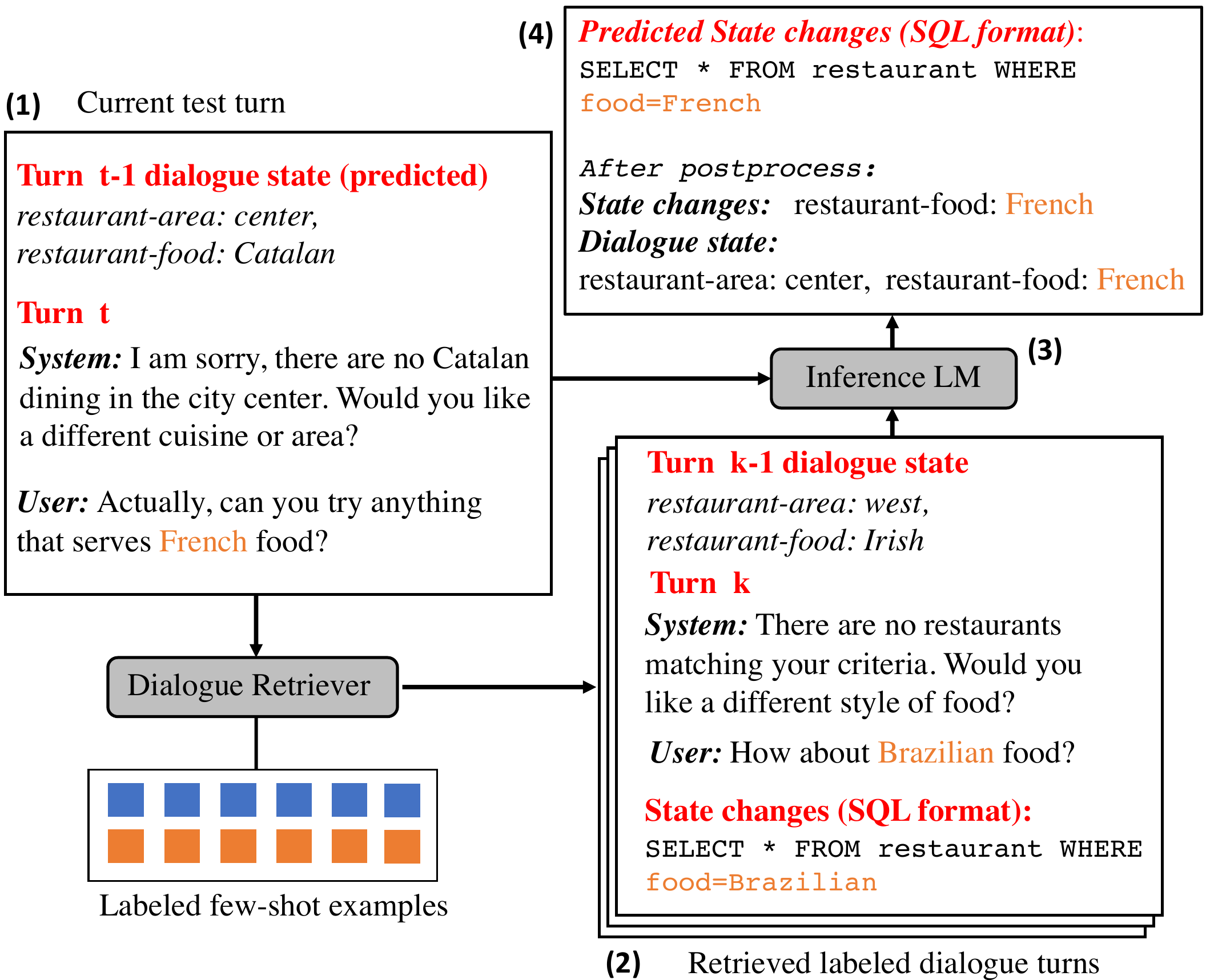}
  \caption{
Illustration of DST task and \NAME approach. 
The task is to track the slot values associated with a user request up to the current turn (dialogue state). 
In few-shot settings,
given a test turn (1), \NAME first retrieves a few most similar turns from the labeled dialogues as examples (2). 
The task schema (not shown in the figure), examples, and the test dialogue turn are concatenated in the prompt to a LM (e.g. GPT3) (3) to produce the current turn dialogue state changes as a SQL query (4).
}
  \vspace{-2mm}
  \label{fig:system}
\end{figure}

\section{Introduction}
\label{introduction}

Dialogue state tracking (DST) is an important module in many task-oriented dialogue systems. 
The goal of this module is to extract users' intentions at each turn in the dialogue as represented in the slot values of a predefined schema. Collecting and annotating turn-level dialogue states is notoriously hard and expensive~\cite{budzianowski2018multiwoz}. Also, in commercial applications, it is common to extend the 
schema and incorporate new domains. Thus, it is important to develop DST learning strategies that are flexible and scalable, in addition to requiring less data.

Previous studies have explored zero/few-shot DST, but with some limitations.
Most few-shot methods are based on finetuning pretrained language models \cite{wu2020improving, li2021zero,su2021multi,shin2022dialogue, lin2021leveraging,xie2022unifiedskg}. 
Such systems are less flexible, since they need to be retrained when new slots or domains are added, and finetuning large LMs is computationally expensive.
Most zero-shot methods have involved domain-transfer approaches \cite{hosseini2020simple,lin2021leveraging,lin2021zero}, which have not yielded good performance.

To address the above challenges, we propose the \NAME model to solve the DST problem with the in-context learning (ICL) paradigm~\cite{brown-gpt3}, in which a large language model makes predictions based on the task instruction and/or examples in the prompt.
In few-shot settings, the prompt contains exemplars that are retrieved from a small set of labeled training data. 
A motivation behind this framework is that it requires no finetuning (i.e., no parameter updates), which
makes systems
flexible 
in that they can handle queries in a new domain via the exemplar retrieval process \textit{without re-training}.
This enables developers to quickly prototype systems in new domains and rapidly leverage new collected data.
ICL has been used successfully in semantic parsing ~\cite{Rajkumar2022EvaluatingTT,pasupat2021controllable,rubin2021learning}, especially in few-shot scenarios. 
However, these studies focus on sentence-level tasks. 
ICL has been explored for DST \cite{madotto2021few,xie2022unifiedskg}, but the performance fell short of pretraining and domain-transfer approaches to few/zero-shot learning. 
DST involves long, two-party dialogue histories with grounding in a structured ontology.
We believe these challenges cause the poor ICL performance on DST tasks in previous work.


To address these challenges, 
we explore in-context learning with three novel contributions. First, we reformulate DST as a text-to-SQL task, including a tabular description of the ontology in the prompt. This is a better match to the knowledge-grounded scenario, and it takes advantage of large language models pretrained with code:  Codex \citep{chen2021evaluating}, GPT-Neo~\citep{gpt-neo}, and CodeGen~\citep{Nijkamp2022ACP}. Second, we use the dialogue state in representing context, rather than the full conversation history, which is more efficient and better suited to domain changes. Lastly, in the few-shot scenario, we propose a new approach to learning a similarity score for selecting in-context examples that is trained to match similarity based on dialogue state changes. The \NAME approach, which incorporates these advances, achieves a new state of the art on MultiWOZ few-shot settings, i.e.\ when using 1--10\% training data. We also substantially improve the zero-shot state of the art by 10--30\% absolute accuracy on each domain.
A further contribution is an extensive analysis demonstrating impact from each innovation.



In summary, our work makes the following contributions:
\begin{itemize}

\item To our knowledge, we are the first to successfully apply in-context learning for DST, building on a text-to-SQL approach.
     
\item To extend in-context learning to dialogues, we introduce an efficient representation for the dialogue history and a new objective for dialogue retriever design.
     
\item Our system achieves a new state of the art on MultiWOZ in zero/few-shot settings.

\item We provide insights into how in-context learning works for dialogue, including the importance of good in-context examples and the LM's ability to generalize beyond examples.
     

\end{itemize}

     


\section{DST System Design}

\subsection{DST Task Framing}
\label{sec:task-framing}
\paragraph{Notation}
A task-oriented dialogue consists of a sequence of utterances alternating between the user and the system, $A_1, U_1, ..., A_T, U_T$, where $A$ and $U$ represent the system and user utterances, respectively.
The task of DST is to predict the dialogue state $y_t$ at each turn $t$,\footnote{For brevity, ``turn'' is used to mean a pair of system and user turns $(A_t, U_t)$, 
associated with state update intervals.}
given the dialogue context $C_t = [A_1, U_1, \cdots, A_t, U_t]$, where
$y_t$ is a set of slot-value pairs: 
\begin{align*}
    y_t = \{(s_i^t, v_i^t); i=1,\ldots , n_t\} .
\end{align*}
The set of possible slots ${s_i}$ is given in a pre-defined schema. 
The schema can contain multiple domains,
where a ``domain'' corresponds to a backend capability such as hotel or restaurant booking. Each domain is associated with a set of slots;
for example, the `hotel' domain has slots `hotel-name', `hotel-price\_range,' `hotel-area', etc. 
Each observed slot is associated with a value, which may have pre-defined categories (e.g., `hotel-price\_range' may be `cheap,' `moderate,' or `expensive') or open (e.g.\ `hotel-name').
We focus on a multi-domain scenario in this work, in which dialogue states may contain slots from multiple domains.  

One popular way to generate the dialogue states for the current turn is finetuning auto-regressive language models \cite{hosseini2020simple,peng2020soloist}. For each turn, the model takes the dialogue context $C_t$
as input, and generates a sequence of slot-value pairs $(s_i, v_i)$ sequentially.
Equivalently, one can generate a sequence of slot-value pair dialogue state changes.

\paragraph{Dialogue states vs.\ state changes}
\label{approach_statechange}


Dialogues can be lengthy and complex, resulting in dialogue states that can include several slots and values, which means that coverage of the possible states is sparse for few-shot learning. However, the dialogue state change from one turn to the next typically involves a small number of slots. For that reason, we use state \emph{changes} at each turn as a label for prediction.
The concept of state changes is illustrated in Figure~\ref{fig:system}.
Possible state changes include slot addition, slot deletion, and slot value change.  
For example, in the current test turn of Figure~\ref{fig:system}, the user asks for Catalan food in turn $t-1$, and changes it to French food in turn $t$.
The state change updates the dialogue state by replacing `Catalan' with `French'. 
Specifically, given the previous turn dialogue state $y_{t-1}$ and the predicted current turn state changes $c_t$, we update the dialogue state by first copying $y_{t-1}$ to $y_t$, and then executing add, delete and change operations according to each slot-value pair $(s_i,v_i)$ in $c_t$.
Our analysis in Section~\ref{analysis_prompting} shows that using state changes leads to substantial improvements.


\paragraph{DST as Text-to-SQL}
Here we propose a new representation for dialogue states: SQL. This is inspired by the fact that dialogue states are used to determine how to query backend databases for the information users need. Our representation follows three rules: (1) each domain 
is defined as a table and each slot is defined as a column; (2) all the slots and values are in the WHERE clause; and (3) for turns with multiple domains, we rename each domain to $d_1, ..., d_m$. 
Using the SQL state representation with a generative LM, DST becomes a Text-to-SQL problem. This approach is facilitated by language models  pretrained with code (Codex and GPT-Neo) as SQL is closer to the code used for pre-training.

\paragraph{Dialogue context representation}
\label{approach_history}

\begin{figure}[t]
    \centering
    \includegraphics[width=\linewidth]{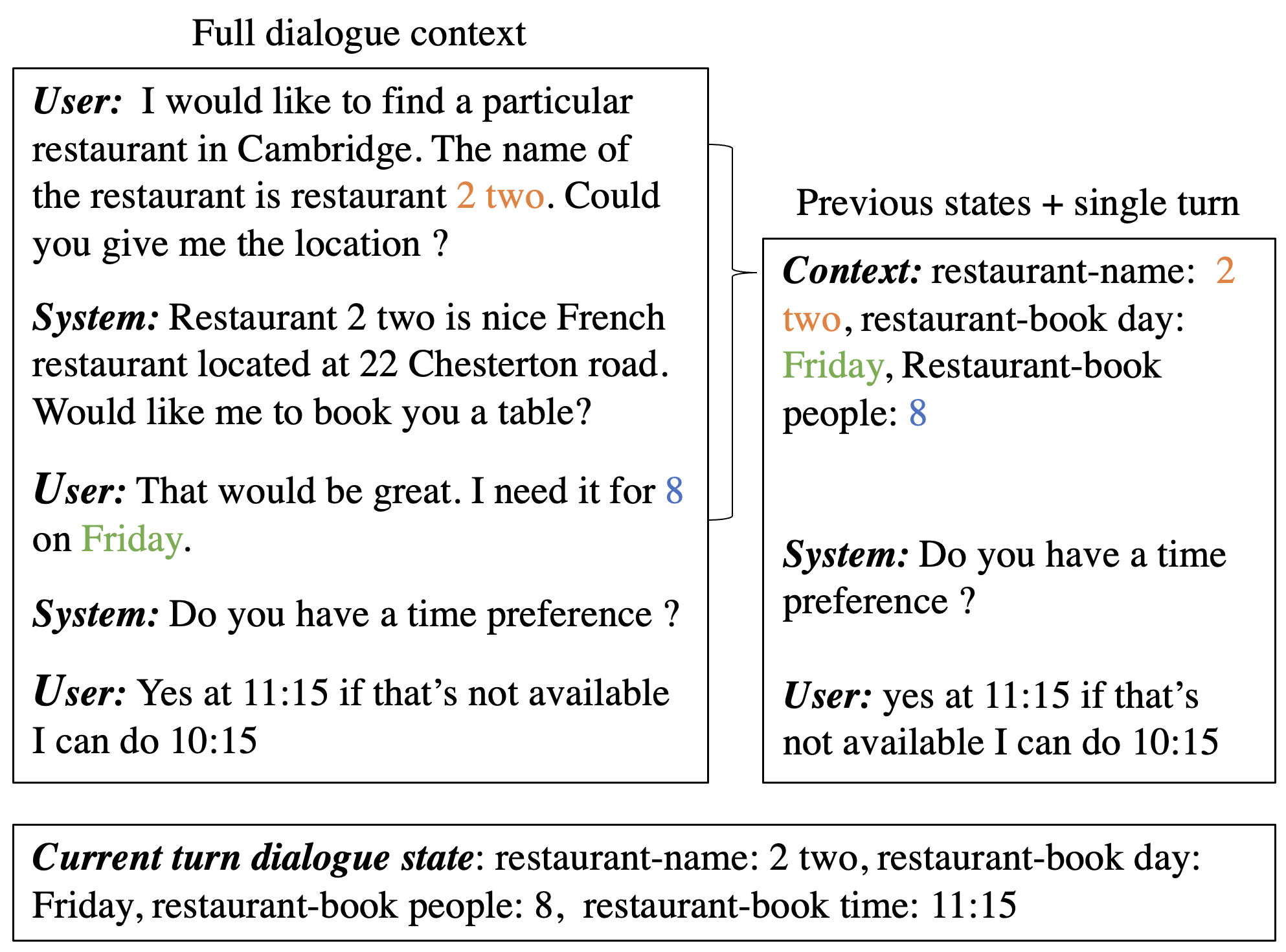}
    \caption{
    Illustration of the dialogue context representation: 
    the full dialogue context $C_{t-1}$ before the current turn is replaced by the associated dialogue state $y_{t-1}$. 
    }
    \label{fig:representation}
    \vspace{-3mm}
\end{figure}

Previous work generally represents dialogue context by concatenating all the system and user utterances $A_1, U_1, \cdots, A_t, U_t$~\cite{lee2021dialogue, lin2021leveraging, peng2020soloist}. 
However, real world dialogues can be lengthy, and there is a length limit for current large language models (2048 tokens for GPT-3, 4096 tokens for Codex). 
It is not practical to represent full dialogue context for multiple exemplars in the prompt. 
A simple solution is to just include the $N$ recent turns in the dialogue history~\cite{lei-etal-2018-sequicity, budzianowski2019hello, wu-etal-2021-dialki}. 
We adopt a new approach that takes advantage of the fact that the dialogue state is a summary of the dialogue history, as shown in Figure~\ref{fig:representation}.
Specifically, 
we represent dialogue context by $[y_{t-1}, A_t, U_t]$, in which $y_{t-1}$ is the accumulated dialogue state after user turn $t-1$.

\subsection{In-Context Learning}


\begin{figure}
  \includegraphics[width=0.48\textwidth]{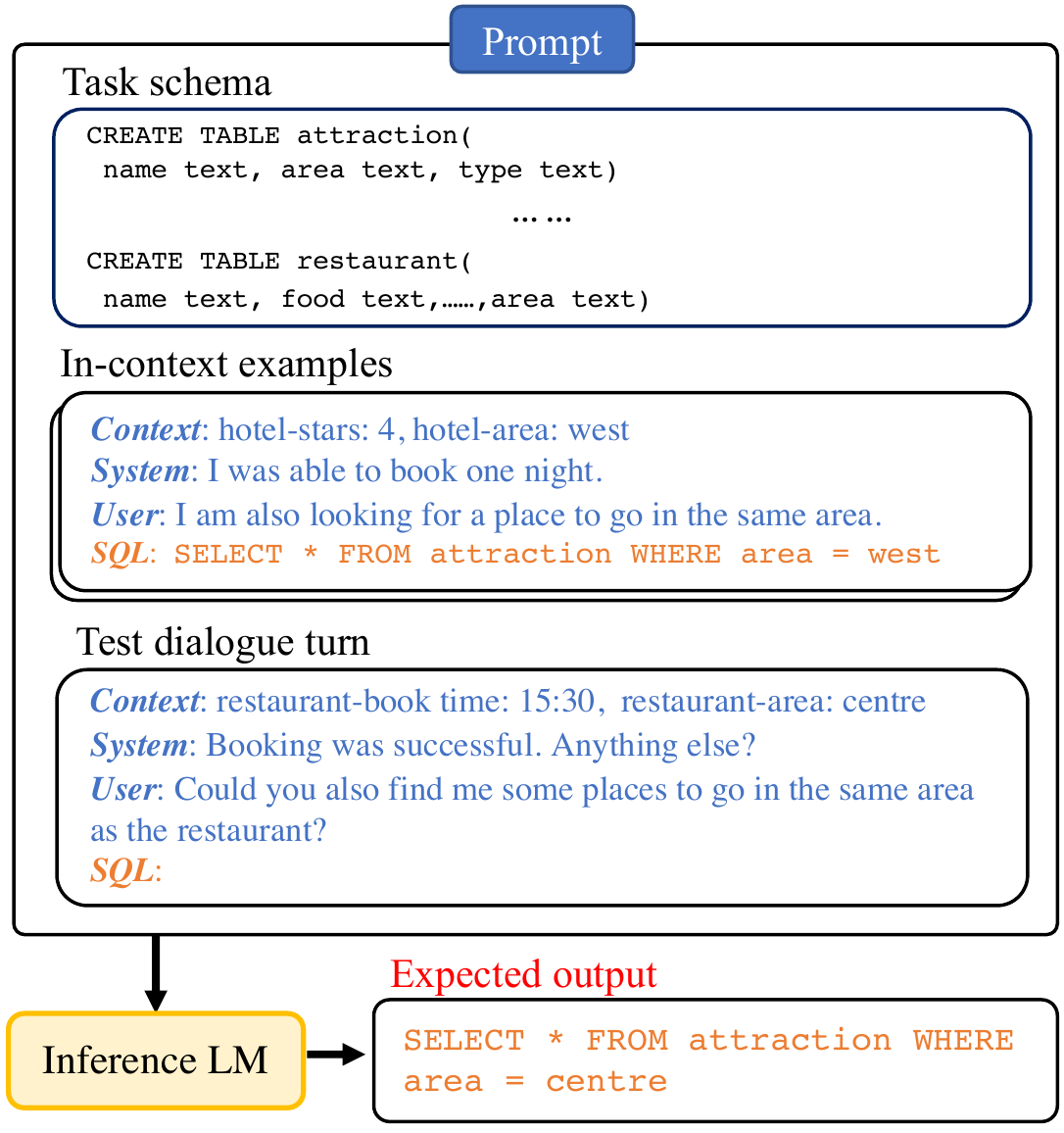}
  \caption{
  The prompt contains the task schema, in-context examples, and the current test dialogue turn.
  }
  \vspace{-3mm}
  \label{fig:prompt}
\end{figure}

\textit{In-context learning} is an alternative to finetuning that keeps pretrained language model parameters fixed~\cite{brown-gpt3}.
The language model takes a prompt $P$ that contains task descriptions, in-context examples, and the test instance as input, and predicts the label by capturing the patterns in the context.
ICL has two advantages over finetuning.
First, it avoids the need for repeated finetuning when the  schema is updated or new examples are added. 
This is particularly important for large models like GPT-3, since finetuning at this scale is extremely expensive. 
Second, by simply adding/removing training examples, in-context learning enables us to quickly manipulate the model's predictions and correct mistakes without re-training.

An overview of our \NAME system is shown in Figure~\ref{fig:system} for the few-shot setting.
The details of our prompt are shown in Figure~\ref{fig:prompt}.
The task description is the schema associated with the task ontology, and a retriever is used to select labeled example turns from the training data.
In the zero-shot setting, there is no retriever.

\begin{figure}[t]
    \centering
    \includegraphics[width=0.9\linewidth]{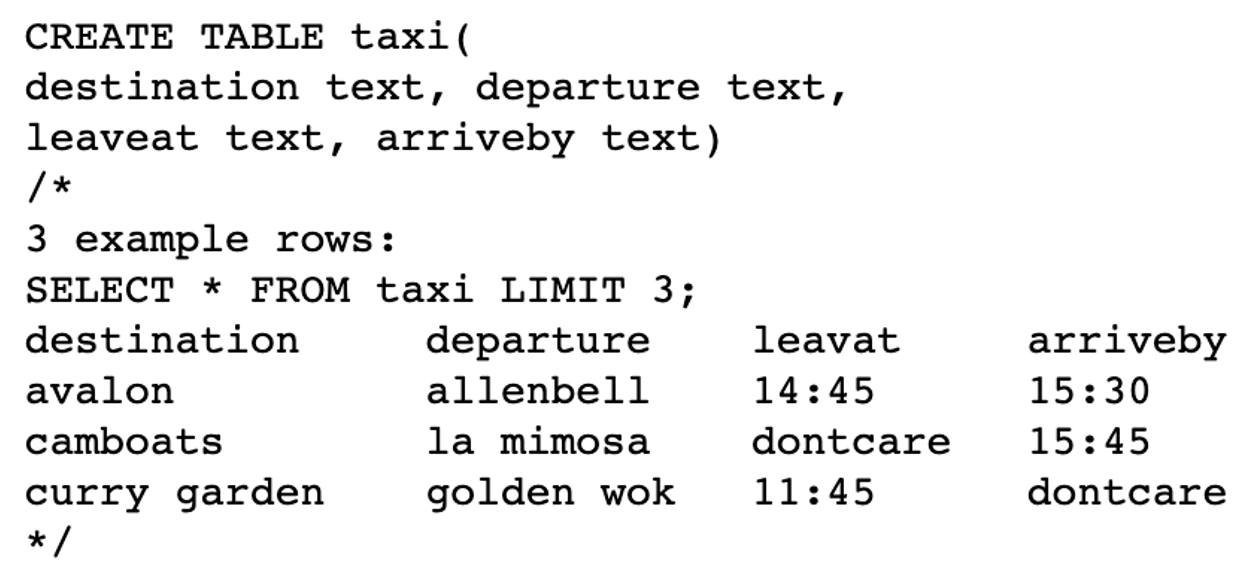}
    \caption{The SQL table for the taxi domain schema. 
    We follow the table prompt from \citet{Rajkumar2022EvaluatingTT}.
    }
    \label{fig:schema}
    \vspace{-3mm}
\end{figure}

\paragraph{Schema prompting}
We use an SQL table for each domain to represent the dialogue schema in the prompt. Each table includes a row of slot names followed by three rows of example values associated with each slot, as illustrated in Figure~\ref{fig:schema}.  
Slots like ``restaurant-name'' or "restaurant-book time" typically have many possible values. Thus, for these slots, we only list a few example values.
In our experiments, we create SQL tables for all domains and concatenate them to be part of our input. 

\paragraph{In-context examples}
In the few-shot scenario,
a retriever takes the dialogue context as input (either $[C_{t-1}, A_t, U_t]$ or $[y_{t-1}, A_t, U_t]$) and retrieves similar example contexts from the labeled training set.
Advantages of using the dialogue state $y_{t-1}$ rather than the full history $C_{t-1}$ are that it is shorter (allowing for more examples) and it leads to a more effective retrieval similarity score.
In the zero-shot scenario, following previous work on zero-shot learning~\cite{wang2022benchmarking}, the in-context example is a formatting example turn.
We call this setting ``zero shot'' because the prompt is fixed by the system developer and does not use any labeled data. 
Prompt examples for the few-shot and zero-shot settings are given in Appendices~\ref{appendix:few shot} and \ref{appendix:zero shot}, respectively.

\subsection{Dialogue Retriever}
\label{approach_retriever}
In few-shot settings, successful prediction of in-context learning relies on the quality of the context examples. 
Usually this is accomplished by semantic retrieval using the testing input as the query. 
Previous studies have explored methods for building sentence-level retrievers~\cite{poesia2022synchromesh,rubin2021learning}. 
Our work goes beyond sentences 
to retrieving dialogue histories (contexts).

We want to retrieve example dialogue contexts that are relevant for the predicted state change of a test sample. 
Formally, suppose $X=\{(e_i,c_i)\}$ is a dataset of dialogue context $e_i$ and corresponding state change $c_i$ pairs. 
Each labeled turn in a dialogue is a candidate. 
Given a test dialogue context $x$, our goal is to retrieve $k$ examples $\{(e_1, c_1), (e_2, c_2), \cdots, (e_k, c_k)\}$ such that $c_1, c_2, \cdots, c_k$ should be similar to the state changes associated with $x$. 

\paragraph{Unsupervised retriever}
One approach is to use a pretrained embedding model with a cosine similarity score as the retriever. This approach does not need any DST data.  
Let $M$ be an embedding model.
For each test example $x$, 
we retrieve the $k$ training examples such that $e_i, e_2, \cdots, e_k$ are the nearest neighbors of $x^*$, given by a similarity score,
\begin{align*}
    \mathit{score}(x,e) = \text{cos}(M(x), M(e)).
\end{align*}
We experiment with RoBERTa~\cite{liu2019roberta} and SBERT~\cite{reimers2019sentence} as embedding models. We also try BM25~\cite{robertson2009probabilistic}, a retriever based on surface text similarity. We find SBERT leads to the best result; therefore, all IC-DST results reported here 
use SBERT as the retriever.

\paragraph{Retriever finetuning}
We also finetune SBERT on the few-shot examples to get a retriever that is better matched to the objective of predicting state changes. We first define the similarity between state changes. Suppose there are two sets of state changes, $c_a=\{(s_1^a, v_1^a), \cdots, (s_m^a, v_m^a)\}$, $c_b=\{(s_1^b, v_1^b), \cdots, (s_n^b, v_n^b)\}$. 
Let $F(set_1, set_2)$ be the average of two $F_1$ scores calculated by using $set_1$ vs. $set_2$ as the target.
Here we use the standard definition $F_1=\frac{2PR}{P+R}$, in which $P$ is precision, and $R$ is recall.
We define the slot similarity as
\begin{align*}
    F_{\mathit{slot}} = F(\{s_1^a, \cdots,s_m^a\}, \{s_1^b, \cdots, s_n^b\}) ,
\end{align*}
and the slot-value pair similarity as
\begin{align*}
    F_{\mathit{slot\mbox{-}value}} = F(&\{(s_1^a, v_1^a), \cdots,(s_m^a, v_m^a)\},\\
    &\{(s_1^b, v_1^b), \cdots, (s_n^b, v_n^b)\}) .
\end{align*}
Then the similarity between $c_a$ and $c_b$ is
\begin{align*}
    s(c_a, c_b) = \frac{1}{2}(F_{\mathit{slot}} + F_{\mathit{slot\mbox{-}value}}) .
\end{align*}

The positive and negative examples for training sample $x_i = (e_i, c_i)$ are identified by computing $s(c_i, c_j)$ for each sample $x_j=(e_j, c_j)$, sorting, and taking the $k$ highest  and lowest scoring samples, respectively.
We finetune the embedding model $M$ with a contrastive loss so that the similarity between a positive example pair is high and the similarity between a negative example pair is low.



\section{Experiments}
\label{experiments}

\begin{table*}[th]
\small
\centering
\begin{tabular}{lrrrr|rrrr}
\toprule
& \multicolumn{4}{c}{MultiWOZ 2.1} & \multicolumn{4}{c}{MultiWOZ 2.4}\\
 Model & 1\% & 5\% & 10\% & 100\% & 1\% & 5\% & 10\% & 100\%\\
 \hline
 \multicolumn{9}{c}{Baselines (finetuned)} \\
 \hline
 TRADE~\cite{wu2020improving} & 12.58 & 31.17 & 36.18 & 46.00 & - & - & - & 55.05\\
 SGPDST~\cite{lee2021dialogue} & 32.11 & 43.14 & 46.92 & 56.66 & - & - & - & - \\
 DS2 - BART~\cite{shin2022dialogue} & 28.25 & 37.71 & 40.29 & 46.86 & 30.55 & 42.53 & 41.73 & 46.14\\
 DS2 - T5~\cite{shin2022dialogue} & 33.76 & 44.20 & 45.38 & 52.32 & 36.76 & 49.89 & 51.05 & 57.85 \\
 \hline
 \multicolumn{9}{c}{In-Context Learning} \\
 \hline
 \NAME GPT-Neo 2.7B & 16.70  &  26.90  & 31.65  & 39.18 & 17.36  & 29.62 & 34.38  & 45.32 \\
 \NAME CodeGen 2.7B & 20.72 &  29.62  & 33.81  & 39.93 & 21.87  & 33.16 & 37.45  & 45.71 \\
 \NAME Codex-davinci & \textbf{43.13} & \textbf{47.08} & \textbf{48.67} & 50.65 & \textbf{48.35}  & \textbf{55.43}  & \textbf{56.88}  & 62.43\\
\bottomrule
\end{tabular}
\caption{
\label{mw2.1results}
Multi-domain JGA evaluated on MultiWoz 2.1 and 2.4 using 1\%, 5\%, 10\%, and 100\% of the training set. 
The average results of 3 runs are reported. 
DS2-T5~\cite{shin2022dialogue} is the previous state-of-the-art few-shot DST model on MultiWOZ. 
}
 \vspace{-3mm}
\end{table*}

\subsection{Datasets}
\label{datasets}

\paragraph{MultiWOZ~\cite{budzianowski2018multiwoz}} is a multi-domain human-human written dialogue dataset that contains over 10K dialogues across 8 domains.
The labels and utterances have been refined in subsequent versions, e.g., 
MultiWOZ 2.1~\cite{eric2019multiwoz}
and MultiWOZ 2.4~\cite{ye2021multiwoz}. MultiWOZ 2.4 is built on top of the 2.1 version and made substantial changes to the validation and testing sets. It can be viewed as a cleaner version of MultiWOZ 2.1 that better reflects model performance. In general, DST results are higher on 2.4 when compared to 2.1.\footnote{Many prior few-shot DST studies report results on MultiWOZ 2.1; we include this older version for direct comparisons.}


\subsection{Baselines}
\label{baselines}


\paragraph{TRADE~\cite{WuTradeDST2019}  }
An encoder-decoder framework is applied to the DST problem, enabling generalization to unseen values and domains. This was the first work to explore cross-domain transfer in DST. Different from IC-DST, TRADE has to make a prediction for each domain and slot pair in separate passes.



\paragraph{SGP-DST~\cite{lee2021dialogue}}
In SGP-DST, schema information is used as a prompt to query a sequence-to-sequence language model (e.g., T5). It achieves SOTA on MultiWOZ 2.2. Similar to TRADE, the value for each domain and slot pair is predicted in a separate pass.

\paragraph{TransferQA~\cite{lin2021zero}}
TransferQA reformulated DST as QA problem. 
It is the state-of-the-art model for zero-shot DST.
The model is pretrained with a large amount of QA data. At inference time, the model predicts slot values by taking synthesized extractive questions as input.

\paragraph{DS2~\cite{shin2022dialogue}}
In DS2, DST is reformulated as a dialogue summarization problem. Sequence-to-sequence language models are trained with synthetic summary templates. The dialogue states can be recovered by reversing the template generation rules. This is by far the strongest few-shot model in the literature, outperforming recent few-shot models like T5-DST~\cite{lin2021leveraging}. However, different from us, this model still requires finetuning on DST labels.



\subsection{Experimental settings}

\paragraph{Few-shot setting} We follow the multi-domain scenario from ~\citet{wu2020improving}, where 1\%, 5\%, and 10\% of training data are sampled as the selection pool. The retriever is fine-tuned on the selection pool and does not see any other DST data. 
    
\paragraph{Zero-shot setting} 
\label{experiment_setting_zero_shot}
There are no labeled examples to retrieve, but a single formatting example turn is included, following previous zero-shot learning work~\cite{wang2022benchmarking}.

\subsection{Experimental Details}

\paragraph{Language models}
\textbf{GPT3}~\cite{brown-gpt3} is a language model with 175B parameters pretrained on a large web corpus. It demonstrates strong zero-shot results on language modeling benchmarks.
Its successor, \textbf{Codex}~\cite{chen2021evaluating}, is pretrained using open-source code from Github.\footnote{\url{https://openai.com/blog/openai-codex/}}
This enables interesting applications such as code completion. 
In our initial studies, as in \cite{Shin2021FewShotSP}, Codex substantially outperforms GPT3; therefore, we use Codex for the following experiments.
In this paper, we use \textbf{Codex-Davinci}.\footnote{In particular, we use code-davinci-002 engine. 
Some papers mention that the model size of Codex-Davinci is 175B, but OpenAI does not officially confirm that.
}
In addition, we report results using \textbf{GPT-Neo} (2.7B) ~\cite{gpt-neo} and \textbf{CodeGen} (2.7B) ~\cite{Nijkamp2022ACP}. They are both pretrained on Pile~\cite{gao2020pile} and open-source code.



\paragraph{Evaluation}
The standard joint goal accuracy (JGA) is used as the evaluation metric.
It treats a prediction as correct only if for every domain all slots exactly match the ground-truth values. To be consistent with prior work~\cite{WuTradeDST2019}, we report all-domain JGA on few-shot settings and per-domain JGA on zero-shot settings. We also report the $F_1$ on slot-value pairs for analysis.

%
\paragraph{\NAME details}
The retriever is initialized with SBERT all-mpnet-v2 (220M). We use AdamW optimizer~\cite{loshchilov2017decoupled} and set $2 \times 10^{-5}$ as learning rate, 1000 as warmup steps.
During retriever finetuning, for each training sample, we first compute similarity based on target labels. 
The positive and negative examples are the top and bottom 5\% samples among the 10\% nearest neighbors of the sample. 
This sampling strategy gives us hard negative examples.
The maximum length for retriever input is 512. 
With the dialogue state context representation, no turn contexts in MultiWOZ exceed this length limit.
For few-shot experiments,
we use 10 context exemplars in Codex experiments and 5 for GPT-Neo due to different length limits.
We set temperature to $0$ to enable greedy argmax sampling during generation.

\section{Results}
\label{result}

\begin{table*}[th]
\small
\centering
\begin{tabular}{lccccc}
\toprule
& attraction & hotel & restaurant & taxi & train\\
 \hline
 \multicolumn{6}{c}{MultiWOZ 2.1}\\
 \hline
SimpleTOD++~\cite{hosseini2020simple} & 28.01 & 17.69 & 15.57 & 59.22 & 27.75 \\
T5DST + description ~\cite{lin2021leveraging} & 33.09 & 21.21 & 21.65 & 64.62 & 35.42 \\
TransferQA ~\cite{lin2021zero} & 31.25 & 22.72 & 26.28 & 61.87 & 36.72 \\
\hline
 \NAME Codex & \textbf{59.97} & \textbf{46.69} & \textbf{57.28} & \textbf{71.35}  & \textbf{49.37}\\
 \hline
 \multicolumn{6}{c}{MultiWOZ 2.4}\\
 \hline
 \NAME Codex & \textbf{62.09} & \textbf{53.18} & \textbf{54.87} & \textbf{71.87}  & \textbf{51.42}\\
\bottomrule
\end{tabular}
\caption{
\label{mw_zeroshot}
Zero-shot per-domain JGA on MultiWOZ 2.1 and 2.4. 
For reference, the multi-domain JGA for IC-DST on MultiWOZ 2.4 is 35.3\%.
}
\vspace{-3mm}
\end{table*}

\paragraph{Few-shot DST on MultiWOZ}
Table~\ref{mw2.1results} shows the result on few-shot settings and full-shot settings of our \NAME compared with several recent baselines 
on MultiWOZ 2.1 and 2.4. 
As discussed in Section~\ref{datasets}, MultiWOZ 2.4 is a clean version of MultiWOZ 2.1 and therefore the performance is better. 
Our system achieves state-of-the-art performance for 1\%, 5\%, and 10\% few-shot learning settings using Codex, outperforming previous works that require model finetuning.  
When given more data as retrieval candidates, our systems improve.
GPT-Neo and Codegen have a similar trend but are generally worse than Codex and other baselines. This suggests that the size of language model matters when deploying ICL.

The prior ICL DST systems did not report on the standard few-shot configurations, so were not included as baselines.  However, our system represents a significant advance over these systems as well, substantially outperforming both UnifiedSKG~\cite{xie2022unifiedskg} (43.1\% vs.\ 23.5\% when using about 80 labeled dialogues) 
and Few-Shot Bot~\cite{madotto2021few} (50.6\% vs.\ 13.9\%) on MultiWOZ 2.1.

\paragraph{Zero-shot DST on MultiWOZ}
Table~\ref{mw_zeroshot} shows the zero-shot DST results on MultiWOZ 2.1 and 2.4. 
Our \NAME outperforms previous results by a large margin. 
In addition, \NAME has the advantage over these approaches that no training is required.
The multi-domain JGA of our \NAME is 35.3\% on MultiWOZ 2.4, which can be compared to 48.35\% for the system using few-shot learning with 1\% training (roughly 80 labeled dialogues).
SimpleTOD++~\cite{hosseini2020simple} and T5DST~\cite{lin2021leveraging} are
trained on four domains on MultiWOZ and tested on the unseen domain. 
In addition, T5DST uses human-written slot descriptions to boost zero-shot performance.
TransferQA~\cite{lin2021zero} does not need any training on DST data.
However, each slot has to be reformatted into a question, and the model is trained on a large amount of QA data.
Our results show the flexibility 
of {\NAME}.
For each new domain or slot added, by updating the SQL tables and adding a demonstration example, the model attains good performance on the new ontology without any training.


\section{Analysis}
\label{analysis}
\vspace{-2mm}

To better understand the effectiveness of our proposed methods, we provide detailed analysis in this section.
All ablation experiments are conducted on 100 random MultiWOZ 2.4 development set dialogues in the 5\% few-shot setting.

\begin{table}[h]\small
\centering

\begin{tabular}{lcr}
\toprule
Dialogue context & Example label & MW 2.4\\
representation &  (retrieval objective) & JGA \\
\midrule
full context & dialogue state &  45.0\\
prev.~state + single turn & dialogue state & 47.9\\
\hline
full context & state changes &  52.0\\
single turn & state changes & 55.8 \\
prev.~state + single turn & state changes & 58.7 \\
\bottomrule
\end{tabular}
\caption{
\label{prompting_method}
Comparison of DST dialogue context representation in prompt examples and different retrieval objectives, with 5\% of the training data and Codex. 
The retrieval objectives correspond to $F_1$ over all slot-value pairs in the dialogue state vs.\ just state changes.
}
\vspace{-5mm}
\end{table}

\paragraph{Representation of dialogue context}
\label{analysis_prompting}
Table~\ref{prompting_method} compares approaches to representing dialogue context in the examples in the prompt, with different retriever fine-tuning objectives. 
For each setting, we train a retriever with the given dialogue context representation and retrieval objective for a fair comparison.
We experiment with representing the dialogue history by:
\textbf{(1)} 
concatenating the whole dialogue history,  
\textbf{(2)} 
only the latest turn (one system utterance and one user utterance), and \textbf{(3)} 
the previous dialogue states and the latest turn.
For both of the retrieval objectives, representing the dialogue context by the previous dialogue state and the current turn gives the best performance. 
There are multiple possible explanations.
First, the previous state is more relevant than the full context for predicting the next-turn dialogue state, especially after a topic shift.
Second, full dialogue contexts are too long. To fit in the length limit of Codex, we sometimes need to truncate examples when using full context, which lowers the system performance, as evidenced by the fact that a single turn outperforms the full context but not the state-based representation.

\begin{figure}[t]
    \centering
    \includegraphics[width=\linewidth]{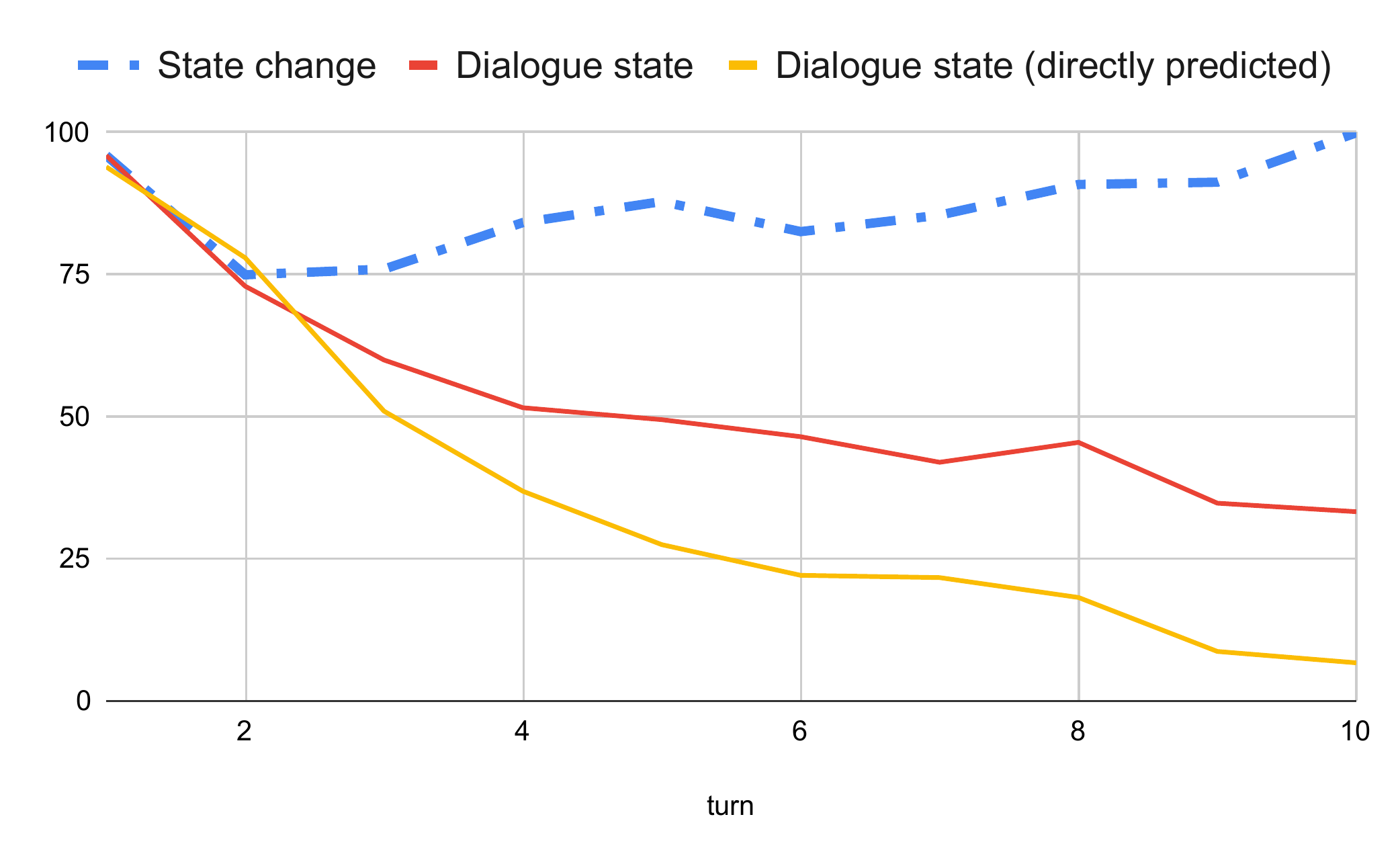}
    \caption{JGA of each turn.  
    The blue line is the JGA of state changes predicted by the system on Table~\ref{prompting_method} row 5 (our \NAME setting).
    The red line is the JGA of dialogue states produced by accumulating predicted state changes.
    The yellow line is the JGA  of dialogue states predicted by the system on Table~\ref{prompting_method} row 2.}
    \label{fig:turn_accuracy}
    \vspace{-4mm}
\end{figure}

\paragraph{Dialogue states vs.\ state changes}

We also explore the benefit of using state change vs.\ full dialogue state labels ($c_t$ vs.\ $y_t$).
Note that example labels also act as the basis for the similarity objective in training the retriever.
%
Table~\ref{prompting_method} shows that using state changes gives substantial improvement.
A likely reason is that state changes contain fewer slots compared with dialogue states, making for an easier prediction task for the LM.
Also, there are many more example turns with the same state changes, compared with examples with the same dialogue states,
making it
easier for the retriever to fetch good examples 
in few-shot scenario. 

We further compare these two kinds of example labels by investigating the JGA on each turn, illustrated in Figure~\ref{fig:turn_accuracy}. 
Because states have increasing numbers of slots with more turns, JGA (red and yellow lines) of the full state decreases for later turns. However, all turns have a relatively small number of state changes, so the state change JGA (blue line) remains high throughout the dialogue.  As a result, the JGA of the full state benefits from using predicted state change updates (red line, Table~\ref{prompting_method} row 5) as compared to predicting the full state (yellow line, Table~\ref{prompting_method} row 2).


\begin{table}[h]\small
\centering

\begin{tabular}{llcccc}
\toprule
&&\multicolumn{2}{c}{Copy} & \multicolumn{2}{c}{\NAME Codex} \\
&& JGA & $F_1$  &  JGA & $F_1$ \\
\hline
Unsuper. &Random & 0.7 & 4.9 & 29.8  & 81.1\\
 & BM25   & 4.9 & 31.8 & 49.0 &  91.0\\
 &SBERT  & 5.9 & 28.7 & 49.2 & 91.7\\
\hline
Super. & \textbf{Ours} & \textbf{16.3} & \textbf{61.4} & \textbf{58.7} & \textbf{93.3}\\
 \hline
 \hline
Oracle & $F_1$-oracle & 24.6 & 76.8 & 82.9  & 98.1\\
\bottomrule
\end{tabular}
\caption{
\label{retriever}
Comparison of different retrievers using 5\% of the training data and Codex. 
The ``copy'' baseline copies the label of  retrieved nearest example as prediction.
The oracle retriever assumes prior knowledge of the gold test instance dialogue state.
}
\vspace{-5mm}
\end{table}

\paragraph{Performance dependence on the retriever}
\label{anaysis_retriever}
Table~\ref{retriever} compares Codex's performance when using different retrievers.
$F_1$ is computed over the predicted and gold full dialogue states.
The ``copy'' baseline makes a prediction by copying the label of the nearest example retrieved by the corresponding retriever in the selection pool.
This baseline gives a direct measure of the quality of the retrieved examples.
First, the result shows that Codex's performance is highly dependent on the quality of the retrievers and their retrieved examples. 
We can also see that $F_1$ of ``copy'' baseline  correlates  with the end task performance (JGA), which supports the use of state-change $F_1$ as a learning objective for 
our similarity measure. 
Our finetuned SBERT retriever obtains substantial improvements over the unsupervised one. 
The oracle retriever fetches the examples that have the highest state change $F_1$,  assuming prior knowledge of test instance gold dialogue states. It shows the upper bound of the retriever performance in this few-shot setting. Note that the state-of-the-art full-shot performance on MultiWOZ 2.4 is 73.62\%~\cite{ye2021slot}. The oracle experiment shows that 
the few-shot ICL may match or outperform the state-of-the-art full-shot model
with further retriever improvements.

Table~\ref{retriever} also shows that the LM is \emph{not} simply copying the example labels. For both evaluation metrics and all the retrievers, the ``copy'' baseline performance is much lower than \NAME, indicating that the inference language model is generalizing over the examples, rather than copying the closest example labels.




\begin{table}[h]\small
\centering

\begin{tabular}{lcc}
\toprule
& Traditional & Text-to-SQL \\
\hline
GPT-Neo 2.7B & 10.7 & 33.2 \\
CodeGen 2.7B & 17.6 & 30.6\\
Codex-Davinci & 49.1 & 58.7 \\
\bottomrule
\end{tabular}
\caption{
Comparison of DST generation target formulation using 5\% training data and Codex.
The traditional format corresponds to the representation in SimpleTOD~\cite{hosseini2020simple}.
}
\label{formulation}
\vspace{-3mm}
\end{table} 

\paragraph{Effect of DST as Text-to-SQL}
Table~\ref{formulation} shows the performance of our ICL framework given different input-output formats. 
We follow SimpleTOD~\cite{hosseini2020simple} as the traditional format to formulate the input-output.
More specifically, the exemplar labels and the generation target are slot-value pairs.
Also, we rewrite the dialogue schema in the same format in the prompt to replace the SQL tables.
The prompt in the setting is shown in Appendix~\ref{appendix:traditional}.
By reformulating DST as a text-to-SQL task, and leveraging large language models pretrained with code, ICL can make better use of the structured knowledge associated with the task schema. 
As shown in Table~\ref{formulation}, GPT-Neo, CodeGEN, and Codex all perform better with text-to-SQL format.
Note that the performance of GPT-Neo with the traditional format is much worse than with the text-to-SQL format, possibly
due to its much smaller model size compared to Codex. It is easier for GPT-Neo to work with SQL, rather than learning a slot value pair format.




\paragraph{Error analysis}

\begin{table*}[th]
    \small
    \centering

    \vspace{-0.5\baselineskip}
    \begin{tabular}{l|p{13cm}}
        \toprule
        \textbf{Error type I} & \textbf{Noisy training example} \\
        \hline
        \textbf{Example} & [user] I need some information about churchill college.\\
        \textbf{Example label} & \textit{attraction-name: churchill college} \\
        \textbf{Test turn} & [user] I am hearing some good things about queens college, can you give me some basic info on them? \\
        \textbf{Prediction} & \textit{attraction-name: queens college} \\
        \textbf{Gold} & \textit{attraction-name: queens college, \textcolor{red}{attraction-type: college}} \\
        \midrule
        

        
         \textbf{Error type II} & \textbf{Retrieval limitations}\\
        \hline
        \textbf{Example} & [system] I found the Scudamores Punting co. and the Cambridge Punter. Which would you prefer? [user] I like the Cambridge Punter better. Can you give the phone number and postcode for them? \\
        \textbf{Example label} & \textit{attraction-name: Cambridge Punter}  \\
        \textbf{Test turn} & [system] There are 2 in the centre of town. Scudamores Punting co., and the Cambridge Punter. Would either of those interest you? [user] could you give me the address for the Cambridge Punter, please? I also need a place to stay , preferably somewhere cheap.\\
        \textbf{Prediction} & \textit{attraction-name: Cambridge Punter} \\
        \textbf{Gold} & \textit{attraction-name: Cambridge Punter, \textcolor{red}{hotel-pricerange: cheap}} \\
        \midrule

        \textbf{Error type III} & \textbf{Failure to generalize from examples}\\
        \hline
        \textbf{Example} & [user] I would like some information on places to stay in Cambridge. I prefer a guesthouse that includes free WiFi, parking does not matter. \\
        \textbf{Example label} & \textit{hotel-internet: yes, hotel-type: guest house, hotel-parking: don't care} \\
        \textbf{Test turn} & [system] What area of town would you like to be in? [user] It doesn't matter. I just want it to be a cheap guesthouse with WiFi included. \\
        \textbf{Prediction} & \textit{hotel-internet: yes, hotel-type: guest house} \\
        \textbf{Gold} & \textit{hotel-internet: yes, hotel-type: guest house, \textcolor{red}{hotel-pricerange: cheap, hotel-area: don't care}} \\
        \bottomrule
    \end{tabular}
        \caption{The most common error types of IC-DST Codex and their corresponding most similar examples. 
        The missed slots in gold state changes are marked in red.}
    \label{tab:common error}
\end{table*}

In examining a subset of \NAME errors, we identified three common types, 
as shown in Table~\ref{tab:common error}.
The first type of error is caused by a \textit{noisy training example}, such as a missing slot. 
ICL is sensitive to noisy training data because the inference LM only sees a few examples in the prompt during prediction. 
%
%
The second type of error is \textit{retrieval limitations}. 
In this case, the retrieved samples are not good exemplars, because they lack some slots that should be predicted in the test instance. This could be due to sparse annotated data (which impacts all few-shot learning methods) or a retriever error.
%
The third type of error is \textit{failure to generalize from examples}, which happens when the model fails to learn from examples. 

\section{Related Work}

\paragraph{Dialogue state tracking}
Many DST systems have been proposed~\cite{WuTradeDST2019,zhang2020task,peng2020soloist,lin2020mintl,lee2021dialogue,zhao2022description,kim2020efficient,heck2020trippy}.
\citet{hosseini2020simple}, \citet{ham2020end} and \citet{peng2020soloist} use naive formats to represent the dialogue states. \citet{cheng2020conversational}, \citet{moradshahi2021contextual}, and \citet{platanios2021value} use hierarchical representations for dialogue states as in semantic parsing tasks.
To reduce the need for labeled data in DST, many approaches are proposed for few-shot DST~\cite{wu2020improving,li2021zero, gao2020machine, lin2021leveraging, campagna2020zero, su2021multi}. 
The state-of-the-art few-shot DST model is~\cite{shin2022dialogue}, in which the authors reformulate DST as a summarization task. We propose to represent the dialogue states as SQL and reformulate DST into a text-to-SQL task.

Most zero-shot methods have involved domain-transfer approaches, including multi-task training with related task-oriented domains \cite{hosseini2020simple,lin2021leveraging} and question-answering datasets \cite{lin2021zero}.
Unfortunately, performance of these systems is quite low.

\paragraph{In-context learning}
ICL of large PLMs like GPT3~\cite{brown-gpt3} has shown increasingly impressive few-shot performance on many NLP tasks.
ICL has obtained success in semantic parsing~\cite{pasupat2021controllable,rubin2021learning,Shin2021FewShotSP}, intent classification~\cite{yu2021few} and other primarily sentence-level tasks~\cite{min2021metaicl}.
To imitate ICL, \citet{pasupat2021controllable} finetuned a T5 model with inputs augmented by retrieved in-context examples.   
\citet{Shin2021FewShotSP} and \citet{Rajkumar2022EvaluatingTT} find that GPT3 or Codex can be generalized to produce different target programs with a few in-context exemplars.
\citet{xie2022unifiedskg} and \citet{madotto2021few} were the first to apply ICL for DST, but their systems underperform other methods.
Future work may consider improving dialogue retrieving methods and task prompting formulation.

\paragraph{Retrieval}
Most current work on ICL focuses on sentences or documents, while our task involves retrieving dialogues.
There are two general kinds of semantic retrievers.
The first is similarity-based retrieval. \citet{poesia2022synchromesh} and \citet{das-etal-2021-case} define a similarity metric between semantic parsing results and use this similarity as the training objective for the retriever. Another approach is LM-score based retrieval. \citet{rubin2021learning} and \citet{Shin2021ConstrainedLM} measure the quality of an example by the probability of a large language model decoding the correct answer. The $k$ highest and lowest quality samples are used as positive and negative samples for the retriever training. The most relevant retrieval studies on dialogue focus on tasks like knowledge identification~\cite{wu-etal-2021-dialki} and response selection~\cite{yuan2019multi, han2021fine}. Their tasks and settings are different from ours.

\vspace{-0.15in}
\section{Conclusion}
\vspace{-0.15in}

We successfully apply in-context learning for dialogue state tracking by introducing a new approach to representing dialogue context, a novel objective for retriever training, and by reformulating DST into a text-to-SQL task.
On MultiWOZ, our system achieves a new state of the art in both few-shot and zero-shot settings. 
Our analyses show that each innovation benefits performance.
We also study in detail the contribution of each design decision.
Future work may apply this in-context learning framework to a wider range of dialogue tasks.

\section{Acknowledgments}

This research was supported in part by funding from Allstate. We thank OpenAI for free access to Codex and Amazon for AWS credits.  



\section{Limitations}
The performance of our framework is highly dependent on the inference language model, which may limit the framework's usage. 
For example, our framework may not work as well on speech recognition transcripts or other languages because of the lack of such data during language model pretraining.
Future work may explore the robustness and generalization ability of in-context learning, for which our \NAME can serve as a test bed.
Also, there is a tradeoff between avoiding the cost of fine tuning with a large language model vs.\ the cost of inference.
Fortunately, thanks to the recent efforts in open-source large models like OPT~\cite{zhang2022opt} and BLOOM~\cite{BLOOM}, and model compression techniques like LLM.int8()~\cite{dettmers2022llmint8}, 
the cost of running large language models has been drastically reduced, and we are optimistic that this trend will continue in the future.
Further, it is possible to leverage zero/few-shot methods as a teacher model to generate “pseudo-labels” for training a system that has a lower inference cost, as in~\citet{ye2022zerogen}.
Future work may investigate more on low-cost approaches to applying in-context learning and large language models.
\bibliography{emnlp2022}
\bibliographystyle{acl_natbib}



\pagebreak
\appendix
\pagebreak
\section{Prompt Examples}
\label{appendix:prompt-example}

\subsection{Prompt in Few-Shot Settings}
\label{appendix:few shot}
Below is the full version of the prompt for few shot settings. Here we are including 5 examples that are retrieved by our finetuned retriever from a 5\% subset of MultiWOZ training set. Notice that the test instance is preceded by ``Example \#6'' but the label needs to be completed by the LM. The completion of Codex is at the end.

\begin{tiny}
\begin{lstlisting}[breaklines]
CREATE TABLE hotel(
  name text,
  pricerange text CHECK (pricerange IN (dontcare, cheap, moderate, expensive)),
  type text CHECK (type IN (hotel, guest house)),
  parking text CHECK (parking IN (dontcare, yes, no)),
  book_stay int,
  book_day text,
  book_people int,
  area text CHECK (area IN (dontcare, centre, east, north, south, west)),
  stars int CHECK (stars IN (dontcare, 0, 1, 2, 3, 4, 5)),
  internet text CHECK (internet IN (dontcare, yes, no))
)
/*
4 example rows:
SELECT * FROM hotel LIMIT 4;
name  pricerange  type  parking book_stay book_day  book_people area  stars internet
a and b guest house moderate  guest house  dontcare  3 friday  5 east  4 yes
ashley hotel  expensive hotel yes 2 thursday  5 north 5 yes
el shaddia guest house  cheap guest house  yes 5 friday  2 centre  dontcare  no
express by holiday inn cambridge  dontcare  guest house yes 3 monday  2 east  dontcare  no
*/

CREATE TABLE train(
  destination text,
  departure text,
  day text,
  book_people int,
  leaveat text,
  arriveby text
)
/*
3 example rows:
SELECT * FROM train LIMIT 3;
destination departure day book_people leaveat arriveby
london kings cross  cambridge monday  6 dontcare 05:51
cambridge stansted airport  dontcare  1 20:24 20:52
peterborough  cambridge saturday  2  12:06  12:56
*/

CREATE TABLE attraction(
  name text,
  area text CHECK (area IN (dontcare, centre, east, north, south, west)),
  type text,
)
/*
4 example rows:
SELECT * FROM attraction LIMIT 4;
name area type
abbey pool and astroturf pitch  centre  swimming pool
adc theatre centre  theatre
all saints church dontcare  architecture
castle galleries  centre  museum
*/

CREATE TABLE restaurant(
  name text,
  food text,
  pricerange text CHECK (pricerange IN (dontcare, cheap, moderate, expensive)),
  area text CHECK (area IN (centre, east, north, south, west)),
  book_time text,
  book_day text,
  book_people int
)
/*
5 example rows:
SELECT * FROM restaurant LIMIT 5;
name  food  pricerange  area  book_time book_day  book_people
pizza hut city centre italian dontcare centre  13:30 wednesday 7
the missing sock  international moderate  east  dontcare dontcare  2
golden wok chinese moderate north 17:11 friday 4
cambridge chop house  dontcare  expensive  center 08:43 monday  5
darrys cookhouse and wine shop  modern european expensive center  11:20 saturday  8
*/

CREATE TABLE taxi(
  destination text,
  departure text,
  leaveat text,
  arriveby text
)
/*
3 example rows:
SELECT * FROM taxi LIMIT 3;
destination departure leaveat arriveby
copper kettle royal spice 14:45 15:30
magdalene college  university arms hotel dontcare  15:45
lovell lodge  da vinci pizzeria 11:45 dontcare
*/

-- Using valid SQLite, answer the following multi-turn conversational questions for the tables provided above.

Example #1
[context] attraction-area: centre, attraction-type: museum, train-departure: cambridge, train-day: friday, train-arrive_by_time: 12:45, train-book people: 6, train-destination: leicester
[system] i recommend castle galleries located at unit su43 , grande arcade , saint andrews street . their phone number is 01223307402 . is there anything else i can help you with ?
Q: [user] excellent , can you give me the postcode ?
SQL: SELECT * FROM attraction WHERE name = castle galleries;


Example #2
[context] attraction-type: museum, restaurant-book day: wednesday, restaurant-book people: 7, restaurant-name: loch fyne, restaurant-book time: 16:30, attraction-area: west
[system] i would suggest cafe jello gallery located at cafe jello gallery , 13 magdalene street . they have free entry .
Q: [user] okay great ! what is their phone number please ?
SQL: SELECT * FROM attraction WHERE name = cafe jello gallery;


Example #3
[context] attraction-area: centre, attraction-type: museum
[system] the broughton house gallery is in the centre , and it has no entrance fee .
Q: [user] may i have the telephone number please ?
SQL: SELECT * FROM attraction WHERE name = broughton house gallery;


Example #4
[context] train-arrive_by_time: 21:30, train-destination: leicester, train-day: thursday, train-departure: cambridge
[system] how many tickets please ?
Q: [user] i do not need to make the reservation now . thank you though . i would like the address for cambridge contemporary art please .
SQL: SELECT * FROM attraction WHERE name = cambridge contemporary art;


Example #5
[context] attraction-area: east
[system] i like the cambridge artworks it s a museum at 5 greens road and it has free admission .
Q: [user] that sounds , good , what is the postcode ?
SQL: SELECT * FROM attraction WHERE name = cambridge artworks;


Example #6
[context] attraction-area: east
[system] how about cambridge artworks ? it s a museum on the east side of town , and they have no entrance fee .
Q: [user] that sounds great . what s their address and postcode ?
SQL: SELECT * FROM
-------------------Prompt Ends here!--------------------
LM completion:  attraction WHERE name = cambridge artworks
\end{lstlisting}
\end{tiny}

\subsection{Prompt in Zero-Shot Settings}
\label{appendix:zero shot}
Below is an example of the prompt in zero-shot settings. 
Some slots are renamed to be easier for model to understand. 
There is one crafted turn for task demonstration, which is marked as ``Example \#1''. 
The ``Example \#2'' is the test instance that needs to be completed.

\begin{tiny}
\begin{lstlisting}
CREATE TABLE hotel(
  name text,
  pricerange text CHECK (pricerange IN (dontcare, cheap, moderate, expensive)),
  type text CHECK (type IN (hotel, guest house)),
  parking text CHECK (parking IN (dontcare, yes, no)),
  book_number_of_days int,
  book_day text,
  book_people int,
  area text CHECK (area IN (dontcare, centre, east, north, south, west)),
  stars int CHECK (stars IN (dontcare, 0, 1, 2, 3, 4, 5)),
  internet text CHECK (internet IN (dontcare, yes, no))
)
/*
4 example rows:
SELECT * FROM hotel LIMIT 4;
name  pricerange  type  parking book_number_of_days book_day  book_people area  stars internet
a and b guest house moderate  guest house  dontcare  3 friday  5 east  4 yes
ashley hotel  expensive hotel yes 2 thursday  5 north 5 yes
el shaddia guest house  cheap guest house  yes 5 friday  2 centre  dontcare  no
express by holiday inn cambridge  dontcare  guest house yes 3 monday  2 east  dontcare  no
*/

CREATE TABLE train(
  destination text,
  departure text,
  day text,
  book_people int,
  depart_time text,
  arrive_by_time text
)
/*
3 example rows:
SELECT * FROM train LIMIT 3;
destination departure day book_people depart_time arrive_by_time
london kings cross  cambridge monday  6 dontcare 05:51
cambridge stansted airport  dontcare  1 20:24 20:52
peterborough  cambridge saturday  2  12:06  12:56
*/

CREATE TABLE attraction(
  name text,
  area text CHECK (area IN (dontcare, centre, east, north, south, west)),
  type text CHECK (type IN (architecture, boat, church, cinema, college, concert hall, entertainment, hotspot, multiple sports, museum, nightclub, park, special, swimming pool, theatre))
)
/*
4 example rows:
SELECT * FROM attraction LIMIT 4;
name area type
abbey pool and astroturf pitch  centre  swimming pool
adc theatre centre  theatre
all saints church dontcare  architecture
castle galleries  centre  museum
*/

CREATE TABLE restaurant(
  name text,
  food_type text,
  pricerange text CHECK (pricerange IN (dontcare, cheap, moderate, expensive)),
  area text CHECK (area IN (centre, east, north, south, west)),
  book_time text,
  book_day text,
  book_people int
)
/*
5 example rows:
SELECT * FROM restaurant LIMIT 5;
name  food_type  pricerange  area  book_time book_day  book_people
pizza hut city centre italian dontcare centre  13:30 wednesday 7
the missing sock  international moderate  east  dontcare dontcare  2
golden wok chinese moderate north 17:11 friday 4
cambridge chop house  dontcare  expensive  center 08:43 monday  5
darrys cookhouse and wine shop  modern european expensive center  11:20 saturday  8
*/

CREATE TABLE taxi(
  destination text,
  departure text,
  depart_time text,
  arrive_by_time text
)
/*
3 example rows:
SELECT * FROM taxi LIMIT 3;
destination departure depart_time arrive_by_time
copper kettle royal spice 14:45 15:30
magdalene college  university arms hotel dontcare  15:45
lovell lodge  da vinci pizzeria 11:45 dontcare
*/

-- Using valid SQLite, answer the following multi-turn conversational questions for the tables provided above.

Example #1
[context]
[system]
Q: [user] i am looking for a guest house to stay in the west. i do not need internet .
SQL: SELECT * FROM hotel WHERE type = guest house AND area = west AND internet = no;

Example #2
[context] 
[system] 
Q: [user] i would like a taxi from saint john s college to pizza hut fen ditton .
SQL: SELECT * FROM
-------------------Prompt Ends here!--------------------
LM completion:  taxi WHERE departure = saint john s college AND destination = pizza hut fen ditton
\end{lstlisting}
\end{tiny}

\subsection{Prompt Example for Traditional Format}
\label{appendix:traditional}

Below is the full version of the traditional format prompt following ~\citet{hosseini2020simple} for few shot settings. The ontology is listed in a similar way to the dialogue states representation. Here we are including 5 examples that are retrieved by our finetuned retriever from a 5\% subset of MultiWOZ training set. Notice that the test instance is preceded by ``Example \#6'' but the label needs to be completed by the LM. The completion of Codex is at the end.

\begin{tiny}
\begin{lstlisting}
hotel-name: a and b guest house, ashley hotel, el shaddia guest house, etc.
hotel-pricerange: dontcare, cheap, moderate, expensive
hotel-type: hotel, guest house
hotel-parking: dontcare, yes, no
hotel-book_stay: 1, 2, 3, etc.
hotel-book_day: monday, tuesday, etc.
hotel-book_people: 1, 2, 3, etc.
hotel-area: dontcare, centre, east, north, south, west
hotel-stars: dontcare, 0, 1, 2, 3, 4, 5
hotel-internet: dontcare, yes, no

train-destination: london kings cross, cambridge, peterborough, etc.
train-departure: cambridge, stansted airport, etc.
train-day: monday, saturday, etc.
train-book_people: 1, 2, 3, etc.
train-leaveat: 20:24, 12:06, etc.
train-arriveby: 05:51, 20:52, etc.

attraction-name: abbey pool and astroturf pitch, adc theatre, all saints church, castle galleries, etc.
attraction-area: dontcare, centre, east, north, south, west
attraction-type: architecture, boat, church, cinema, college, concert hall, entertainment, hotspot, multiple sports, museum, nightclub, park, special, swimming pool, theatre

restaurant-name: pizza hut city centre, the missing sock, golden wok, cambridge chop house, darrys cookhouse and wine shop, etc.
restaurant-food: italian, international, chinese, dontcare, modern european, etc.
restaurant-pricerange: dontcare, cheap, moderate, expensive
restaurant-area: centre, east, north, south, west
restaurant-book_time: 13:30, 17:11, etc.
restaurant-book_day: wednesday, friday, etc.
restaurant-book_people: 1, 2, 3, etc.

taxi-destination: copper kettle, magdalene college, lovell lodge
taxi-departure: royal spice, university arms hotel, da vinci pizzeria
taxi-leaveat: 14:45, 11:15, etc.
taxi-arriveby: 15:30, 12:45, etc.

-- answer the following multi-turn conversational questions for the ontology provided above.

Example #1
[context] 
[system] 
Q: [user] find me a restaurant that serves belgian food in the centre
A: restaurant-area: centre, restaurant-food: belgian;


Example #2
[context] 
[system] 
Q: [user] i need a place to dine on indian food . centre of the town please .
A: restaurant-area: centre, restaurant-food: indian;


Example #3
[context] restaurant-food: italian, restaurant-area: centre
[system] sure , did you want someone in a certain price range ?
Q: [user] no , price does not matter .
A: restaurant-pricerange: dontcare;


Example #4
[context] 
[system] 
Q: [user] hello , i am looking for a venetian restaurant in the centre of town .
A: restaurant-area: centre, restaurant-food: venetian;


Example #5
[context] 
[system] 
Q: [user] hello ! i would like to get some italian food , somewhere in the center of town .
A: restaurant-food: italian, restaurant-area: centre;


Example #6
[context] 
[system] 
Q: [user] i am looking for a italian restaurant centre .
A:
-----------------------Prompt Ends here!--------------------
LM completion:  restaurant-area: centre, restaurant-food: italian
\end{lstlisting}
\end{tiny}

\section{Details on Example Errors}
\label{appendix:errors}

This section provides a more detailed description of errors described in Table~\ref{tab:common error}.

The first type of error is caused by a \textit{noisy training example}. In this example, there is a missing slot `attraction-type' in the example label. 

The second type of error is \textit{unable to retrieve good examples}. 
In this example, the inference LM is mimicing the example, only predicting slot `attraction-name' and missing the slot `hotel-pricerange'. 

The third type of error is \textit{fail to generalize from examples}. In this example, the in-context example contains the notion of slot value `don't care'. However, the LM fails to generalize this notion and misses the slot value pair `hotel-area: don't care'.

\section{Computing Details}
For zero-shot inference and few-shot in-context learning, we do not need any training. We use one NVIDIA A10G graphics card for GPT-Neo and CodeGen models. Inference of one turn takes about 2 seconds for both models. For Codex, we use the OpenAI API. On average each turn takes about 4 seconds for inference.

For retriever finetuning, we train all our retrievers on one single NVIDIA A10G graphics card. 
For 5\% few-shot scenario, each epoch of training takes 10 minutes on average with our given hyperparameters, and we train for 10 epochs. For 100\% full-shot scenario, we train for 2 epochs, which takes 6 hours. The training time is linear to the data size.

Only retriever finetuning requires hyperparameter tuning. Hyperparameters are given in our code, and the best configuration is provided in the experiment section. 
They are chosen by manual tuning.
The development evaluation metrics is the average $F_1$ score of retrieved examples, which has been described in the paper.
For learning rates, we experimented over $1 \times 10^{-5}$ and $2 \times 10^{-5}$.
For contrastive learning, we scale the number of positive/negative examples for each data point linearly with the selection pool size.
When using 1\% of labeled data, the number of positive/negative examples is 2. For 5\% of data, the number is 10.
We also experimented with doubling the number of positive/negative examples.
All few-shot test numbers are the average of three evaluation runs, and all retrievers are trained just once with each hyperparameter configuration.

\section{Dataset Details}
MultiWOZ~\cite{budzianowski2018multiwoz} is an English multi-domain task-oriented dialogue dataset. 
It contains 7 different domains.
Following the previous work~\cite{WuTradeDST2019}, we use 5 domains: hotel, taxi, attraction, restaurant, and train. 
There are 8438 dialogues in the training set, and 1000 dialogues in the dev and test set.
On average, there are 13.46 turns per dialogue and 13.13 tokens per turn.
For preprocessing, we use the scripts of~\citet{ye2021multiwoz}.
This script mainly fixes typos and standardizes the formatting.
All data are downloadable from ~\citet{ye2021multiwoz}.

\begin{table*}[h]
\centering

\begin{tabular}{lcccccc}
\toprule
& \multicolumn{3}{c}{MultiWOZ 2.1} & \multicolumn{3}{c}{MultiWOZ 2.4} \\
& 1\% & 5\% & 10\%  & 1\% & 5\% & 10\% \\
\hline
 \multicolumn{7}{c}{GPT-Neo} \\
\hline
Run 1 & 18.44  & 26.24 &  32.10 & 19.29  & 28.79 &  35.22\\
Run 2 & 14.32  & 27.63 & 31.62 & 14.32  & 30.78 & 33.98\\
Run 3 & 17.35  & 26.83 & 31.23 & 18.46  & 29.30 & 33.93\\
\hline
Mean &  16.70 & 26.90 & 31.65 &  17.36 & 29.63 & 34.38\\
(Stdev) & (2.13) & (0.70) & (0.44) & (2.66) & (1.03) & (0.73)\\
\hline
 \multicolumn{7}{c}{CodeGen} \\
\hline
Run 1 & 22.11  & 29.33 &  33.70  & 23.25  & 33.03 &  37.83\\
Run 2 & 17.78  & 30.05 & 34.08 & 19.22  & 33.36 & 37.43\\
Run 3 & 22.27  & 29.47 & 33.66 & 23.13  & 33.08 & 37.08\\
\hline
Mean &  20.72 & 29.62 & 33.81 &  21.87 & 33.16 & 37.45 \\
(Stdev) & (2.55) & (0.38) & (0.23) & (2.29) & (0.18) & (0.38)\\
\hline
 \multicolumn{7}{c}{Codex} \\
\hline
Run 1 & 44.25  & 47.52 &  48.34 & 49.92  & 56.27 &  56.62\\
Run 2 & 41.99  & 47.18 & 49.04  & 47.31  & 54.33 & 56.65\\
Run 3 & 43.16  & 46.55 & 48.64 & 47.83  & 55.69 & 57.37\\
\hline
Mean &  43.13 & 47.08 & 48.67  &  48.35 & 55.43 & 56.88 \\
(Stdev) & (1.13) & (0.49) & (0.35) & (1.38) & (1.00) & (0.42)\\

\bottomrule
\end{tabular}
\caption{
Detailed test JGA in few-shot settings of MultiWOZ 2.1 and MultiWOZ 2.4.
}
\label{mw_details}
\vspace{-3mm}
\end{table*}

\section{Detailed Results}
Table~\ref{mw_details} shows each run's JGA on MultiWOZ 2.1 and 2.4.

\end{document}